\documentclass[letterpaper, 10 pt, conference]{ieeeconf}  

\IEEEoverridecommandlockouts                              

\usepackage{float}
\usepackage[labelformat=simple]{subcaption}

\usepackage{fixltx2e}
\usepackage{graphicx}
\usepackage{multirow}
\usepackage{hyperref}

\title{\LARGE \bf
Computational ergonomics for task delegation in Human-Robot Collaboration: spatiotemporal adaptation of the robot to the human through contactless gesture recognition
}

\author{Brenda Elizabeth Olivas-Padilla$^{1}$, Dimitris Papanagiotou$^{1}$, Gavriela Senteri$^{1}$,
\\
Sotiris Manitsaris$^{1}$, and Alina Glushkova$^{1}$

\thanks{$^{1}$Authors are with the Centre for Robotics, MINES Paris, PSL Université, 75006 Paris, France.
        {\tt\small B.O.(brenda.olivas@mines-paristech.fr);}
        {\tt\small D.P.(dimitris.papanagiotou@mines-paristech.fr);}
        {\tt\small G.S.(gavriela.senteri@mines-paristech.fr);}
        {\tt\small S.M.(sotiris.manitsaris@mines-paristech.fr);}
        {\tt\small A.G.(alina.glushkova@mines-paristech.fr);}}}

\begin{document}

\maketitle
\thispagestyle{empty}
\pagestyle{empty}

\begin{abstract}
The high prevalence of work-related musculoskeletal disorders (WMSDs) could be addressed by optimizing Human-Robot Collaboration (HRC) frameworks for manufacturing applications. In this context, this paper proposes two hypotheses for ergonomically effective task delegation and HRC. The first hypothesis states that it is possible to quantify ergonomically professional tasks using motion data from a reduced set of sensors. Then, the most dangerous tasks can be delegated to a collaborative robot. The second hypothesis is that by including gesture recognition and spatial adaptation, the ergonomics of an HRC scenario can be improved by avoiding needless motions that could expose operators to ergonomic risks and by lowering the physical effort required of operators. An HRC scenario for a television manufacturing process is optimized to test both hypotheses. For the ergonomic evaluation, motion primitives with known ergonomic risks were modeled for their detection in professional tasks and to estimate a risk score based on the European Assembly Worksheet (EAWS). A Deep Learning gesture recognition module trained with egocentric television assembly data was used to complement the collaboration between the human operator and the robot. Additionally, a skeleton-tracking algorithm provided the robot with information about the operator's pose, allowing it to spatially adapt its motion to the operator's anthropometrics. Three experiments were conducted to determine the effect of gesture recognition and spatial adaptation on the operator's range of motion. The rate of spatial adaptation was used as a key performance indicator (KPI), and a new KPI for measuring the reduction in the operator's motion is presented in this paper.
\end{abstract}
\footnote{This work has been submitted to the IEEE for possible publication. Copyright may be transferred without notice, after which this version may no longer be accessible.}
\section{Introduction}
Industry 4.0 has resulted in a rise in research in the field of Human-Robot Collaboration (HRC). As a result, robotic agents are being integrated into the work routine, not to take the position of human operators but to assist them in accomplishing complicated and physically demanding tasks. By integrating collaborative robots properly, operators may be able to avoid developing work-related musculoskeletal disorders (WMSDs). WMSDs are a significant concern in the industry, constituting for the majority of work-related health problems in Europe \cite{JandeKok2019}. These are caused by the repeated performance of difficult and repetitive operations that frequently demand operators to push themselves beyond their normal physical limitations. 

Task delegation and coordination between operators and robots must be optimized while ergonomic aspects are considered to maximize ergonomics and production efficiency in industrial co-production cells. With this in mind, two hypotheses for ergonomically effective task delegation and HRC are proposed in this paper. The first hypothesis (H1) is that by utilizing motion capture (MoCap) data, operators' postures and movements can be accurately measured, allowing for a more thorough ergonomic analysis of their actions. The ability to record accurate measurements for ergonomic analysis is essential as it provides quantitative measures of operators' performance. In this work, the ergonomic analysis method developed is based on segmenting the professional tasks into motion primitives with known ergonomic risk levels. The tasks are initially recorded using a minimized set of inertial sensors to facilitate the implementation of the proposed framework in industrial settings. After evaluation, the collaborative robot is then given the tasks from which the most dangerous motions are detected, while the operators are given the ergonomically safe or supervisory control tasks.

The second hypothesis (H2) is that human gesture recognition and spatial adaptation enable a more natural HRC by performing only motions that are convenient for the operators and avoiding unnecessary movements that may expose them to ergonomic risks. Human gesture recognition is one of the methods used to achieve contactless communication between a robot and a human operator. It can be defined as the process of transforming movements into a form that a machine can easily interpret. On the other hand, spatial adaptation is the process through which the robot adjusts its movements according to the anthropometric characteristics of each operator. By adding spatial adaptation in the HRC, the robot can assist operators in reducing their range of motion and, consequently, the physical effort necessary to fulfill their professional duties.

\begin{figure*}[thpb]
    \vspace{10pt}
      \centering
      \includegraphics[width=0.8\textwidth]{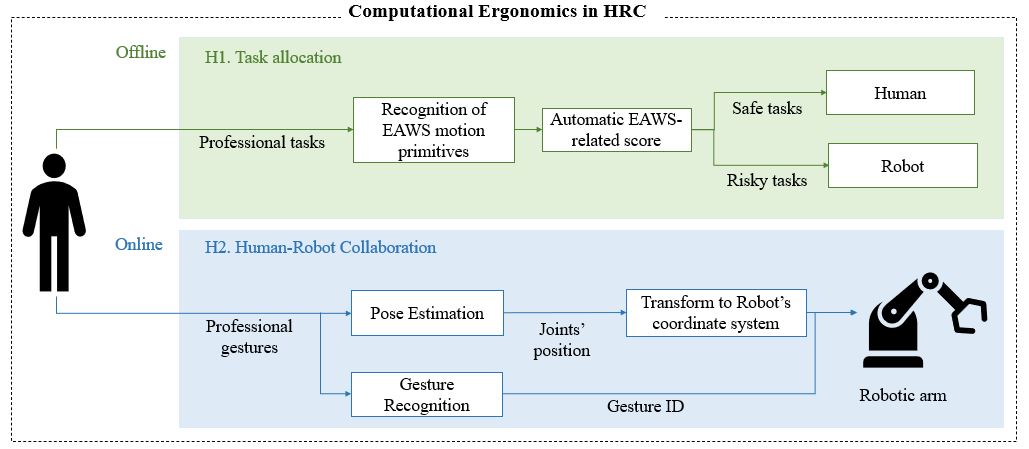}
      \caption{Pipeline for ergonomically optimizing industrial co-production cells with HRC}
      \label{fig:Method}
   \end{figure*}

Fig. \ref{fig:Method} illustrates the pipeline for ergonomically optimizing an industrial HRC scenario. The first hypothesis is tested using real professional tasks recorded in a co-production cell for television manufacturing. Next, the proposed HRC is evaluated using two key performance indicators (KPI) to test the second hypothesis. Section \ref{SoA} discusses the current state of the art in ergonomic analysis using MoCap data and HRC frameworks. Following that, Section \ref{ergoA} explains the methodology for ergonomically evaluating professional tasks and the results obtained. Section \ref{HRCsec} details the HRC's framework and the KPI results. Finally, Section \ref{concF} presents the conclusion and future work.

\section{State of the Art} \label{SoA}
As mentioned earlier, the activities performed by manual laborers in the industrial sector are becoming more challenging and complex in order to meet market demands within certain time limits, job specifications, and budget constraints. Operators must go beyond their natural physical limitations to undertake repetitive jobs for long periods of time in order to complete the tasks required of them. Being subjected to such constant physical strain leads to work-related musculoskeletal disorders (WMSDs). Ergonomists have developed a variety of methods for evaluating work-related tasks. The methods based their analysis on theoretical knowledge of human physical limitations and abilities indicated by known standards (e.g., ISO 11226:2000 and EN 1005-4). Some of the most popular methods are the Rapid Upper Limb Assessment (RULA) \cite{Lynn1993}, Rapid Entire Body Assessment (REBA) \cite{McAtamney2004}, European Assembly Worksheet (EAWS) \cite{Schaub2013}, and Ovako Working Posture Analysing System (OWAS) \cite{Karhu1977}. To implement these methodologies, the ergonomist observes an operator executing the task under evaluation and annotates various body part postures on a worksheet. The ergonomic score of the task is then calculated using these annotations. This method of scoring determines which tasks should be changed for better ergonomics. However, because these approaches rely on the ergonomist's perception and experience, scoring can be subjective and have a lot of inter-variability. Alternative sensor-based ergonomic evaluation approaches are now being developed by researchers. Optical and inertial-based motion capture systems have frequently been used to extract upper body posture for ergonomic evaluation \cite{Manghisi2017, Plantard2016, Shafti2019}. Using inertial measurement units, Vignais et al. created a real-time ergonomic assessment based on RULA (IMUs). Similarly, Yan \cite{Yan2017} used inertial sensors to track the torso inclination of construction workers for ergonomic monitoring.

Collaborative robots are becoming increasingly common in industrial settings as an automated solution for making workplaces more ergonomic, cost-effective, and flexible. As a result, early research focused on creating HRC frameworks that physically couple humans and robots. For example, some applications are for co-carrying \cite{Agravante2014, Yu2021, Sirintuna2022} and co-manipulation \cite{Peternel2017, Wu2020, Al-yacoub2021}. However, the majority of existing real-time HRC frameworks are developed around the concept of human avoidance to assure safety, i.e., preventing unintentional accidents between humans and robots carrying sharp or heavy industrial goods \cite{Liu2019, Sharkawy2020, El-Shamouty2020}. Further research is needed to determine how ergonomics should be considered when a robot collaborates with a human. For instance, when asked to perform the same action, different users of the same setup have different anthropometric characteristics and behaviors. These distinctions must be taken into account in HRC frameworks.

Moreover, meaningful key performance indicators (KPIs) that evaluate either the operators or the robot are also required to measure the performance and interaction of these adaptive HRC. Previous research had offered metrics that primarily assessed the collaborative robot's performance \cite{Goodrich2003, Steinfeld2006}. Therefore, new KPIs must be developed that take into consideration human factors to evaluate the job quality of operators in the HRC, and not only its productivity performance.

For designing ergonomic HRC scenarios, previous studies used biomechanical simulations to compute ergonomic metrics (posture, physical effort, and energy spent during the task) \cite{Maurice2014, Kim18, Marin18}. However, the main downside of these approaches is that they are hard to incorporate into industrial applications that demand rapid reconfigurability. This is because the human ergonomic analysis in these studies is done offline and in a laboratory. In most cases, for an accurate performance evaluation, it is required to use optical motion capture technology, which is only available in specialized laboratories, or the measurements of numerous sensors distributed throughout the operators' bodies. The recordings inside laboratories can cause inaccurate measures since they lack authenticity and are not real workplace scenarios. After simulating professional tasks, an offline biomechanical analysis is done to compute the ergonomic metrics for the workstation redesign. Consequently, methodologies that employ technologies that are simple to implement in real-world scenarios and that can quickly and effectively estimate the ergonomic risk level of any representative set of manipulation actions performed in the industry are still needed.

To summarize, highly adaptable and rapidly reconfigurable frameworks with real-time data processing capabilities are necessary to handle the industry's ergonomics challenges while simultaneously ensuring productivity. This paper expands on the authors' prior work on adaptive HRC \cite{Papanagiotou2021}, and automatic ergonomic monitoring \cite{Olivas-Padilla2019, Olivas-Padilla2020,Olivas-Padilla2021}, with the goal of creating a unified ergonomic and reconfigurable HRC framework.

\section{Automatic ergonomic evaluation of professional tasks} \label{ergoA}
This section describes the task delegation methodology followed to improve the ergonomics of a real-world workplace scenario.

\subsection{Motion capturing of the television assembly scenario}
For the initial study and recording of tasks performed on a television (TV) production line, it was used the BioMed bundle motion capture system from Nansense Inc.\footnote{Baranger Studios, Los Angeles, CA, USA}. This system consisted of a full-body suit composed of 52 IMUs strategically placed throughout the body and hands. At a rate of 90 frames per second, the IMUs were used to calculate the Euler joint angles of body segments on the articulated spine chain, shoulders, limbs, and fingertips.

\begin{figure}[]
\centering
{\captionsetup{position=bottom,justification=centering}
	\begin{subfigure}{0.4\linewidth}
		\centering
		\includegraphics[width=0.8\textwidth, height=25mm]{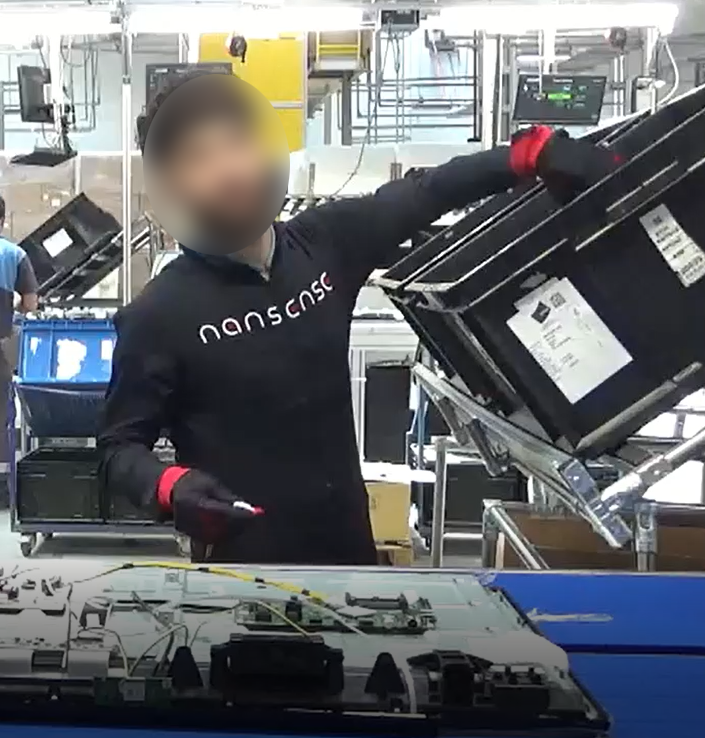}
		\caption{}\label{fig:GV2a}
	\end{subfigure}
	\begin{subfigure}{0.4\linewidth}
		\centering
		\includegraphics[width=0.8\textwidth, height=25mm]{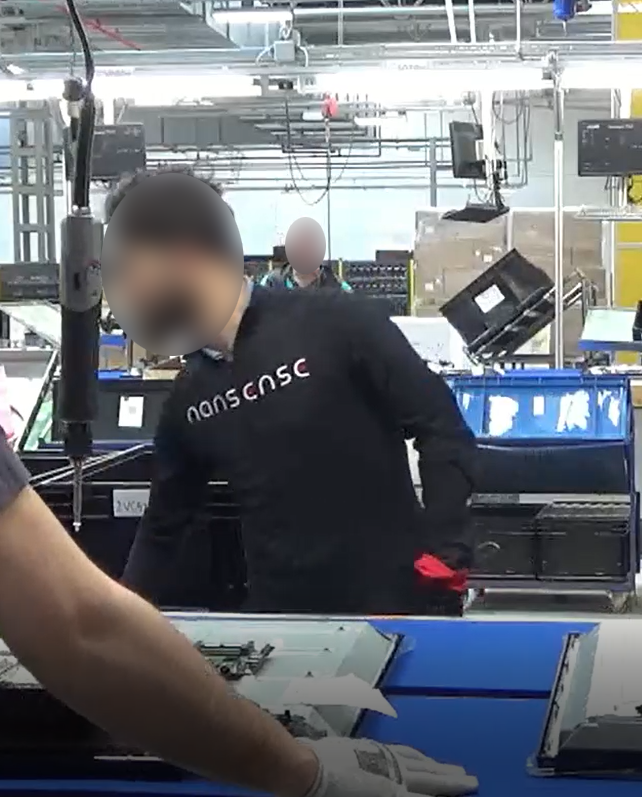}
		\caption{}\label{fig:GV2b}
	\end{subfigure}
	\begin{subfigure}{0.4\linewidth}
		\centering
		\includegraphics[width=0.8\textwidth, height=25mm]{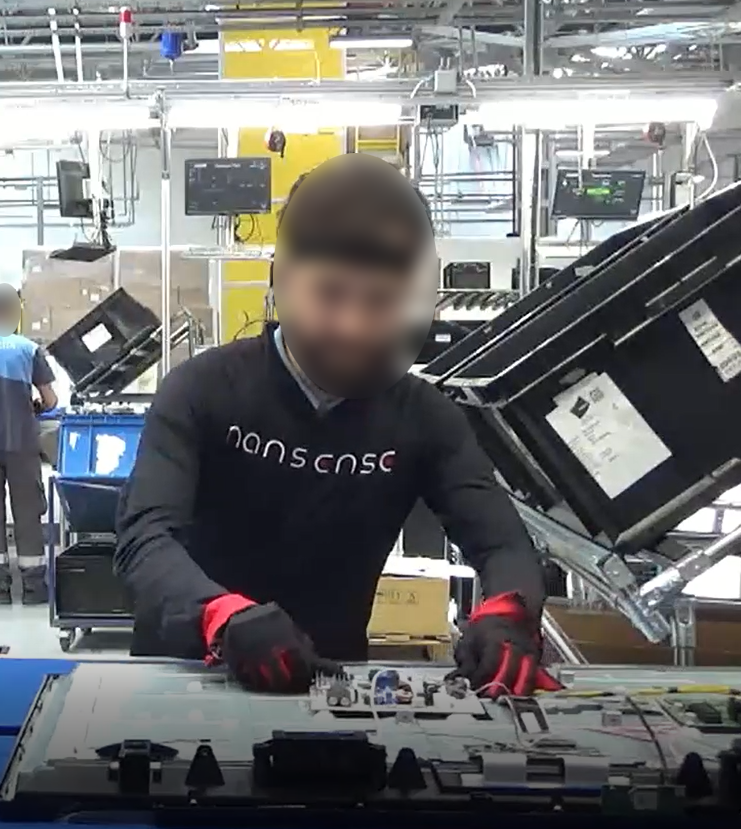}
		\caption{}\label{fig:GV2c}
	\end{subfigure}
	\begin{subfigure}{0.4\linewidth}
		\centering
		\includegraphics[width=0.8\textwidth, height=25mm]{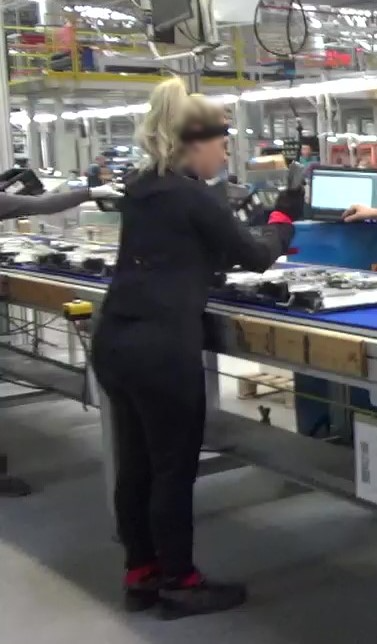}
		\caption{}\label{fig:GV2d}
	\end{subfigure}}
	\caption{Professional tasks for TV assembly. (\textbf{a}) T\textsubscript{1}: Grab the circuit card from a container; (\textbf{b}) T\textsubscript{2}: Take a wire from a container; (\textbf{c}) T\textsubscript{3}: Connect the circuit card and wire and place them on the TV frame; (\textbf{d}) T\textsubscript{4}: Drilling circuit cards to the TV frame.}
	\label{fig:GV2}
\end{figure}

Two healthy adults, one male and one female, wore the IMU-based suit during the MoCap recording session. The recording took place on a real production line over the course of an eight-hour shift, with the operators placing electronic circuit cards into the frame of the televisions. The entire assembly procedure can be broken down into four main tasks, each illustrated in Fig. \ref{fig:GV2}. The first task is grabbing a circuit card from a container (T\textsubscript{1}); the second is taking a wire from a second container (T\textsubscript{2}); the third involves connecting the card and wire and placing them on the TV frame (T\textsubscript{3}); the fourth task corresponds to drilling the circuit cards on the TV frame (T\textsubscript{4}). The whole assembly was recorded 108 times. Following the recordings, the MoCap data was offline pre-processed before analysis. Low pass filtering was used to reduce noise, and the common zero velocity update algorithm was used to eliminate drift caused by electromagnetic interference.

\subsection{Modeling of motion primitives for ergonomic evaluation} 
The methodology described in a previous study \cite{Olivas-Padilla2020} was followed to detect dangerous motions performed in the television assembly tasks. In this study, HMMs were trained to recognize motion primitives with varying ergonomic risk levels according to EAWS. The ergonomic score for each task was then determined based on the motions detected.

For the generation of the data set with the motion primitives, MoCap recordings were made in a laboratory. The neutral environment sessions involved ten healthy individuals, three females and seven males. None of them suffered from any musculoskeletal injury. The protocol contained 28 different motion primitives, each recorded three times. The motions can be classified into three broad categories: those performed standing, those performed while seated in a chair, and those performed while kneeling. The small movements progressed from comfortable to increasingly uncomfortable postures (rising arms, bending forward, rotating the torso). Each motion was assigned an EAWS risk score, which ranges from 0.5 to 26.5. The higher the ergonomic risk score, the greater the risk.

Only motion primitives where subjects were standing were used to train the HMMs, as operators in the television production line are always required to be standing. Thus, HMMs were trained to recognize only 14 motion primitives with different ergonomic risk levels. Additionally, in a second prior study, where these motion primitives were analyzed by modeling their spatiotemporal dynamics through a Gestural Operational Model (GOM) \cite{Olivas-Padilla2021}, it was discovered that with only five inertial sensors, high recognition accuracy can be achieved. These sensors were placed on the lumbar spine, left upper arm, right shoulder, right upper leg, and left forearm. Therefore, the 14 HMMs were trained using joint angle sequences obtained exclusively from these five sensors and using the same HMM topology and number of states recommended by the study.

\subsection{Automatic postural evaluation of TV assembly motions} 
To conduct the ergonomic evaluation, the recordings of the professional tasks were first segmented into windows of similar duration as the motion primitives (approximately five seconds), then provided to the 14 pre-trained HMMs. The resulting likelihoods were used to determine the detected motion primitive. All the motion primitives recognized throughout each task were annotated and used to compute each task's mode and average ergonomic score. The ergonomic score was calculated using the equations in \cite{Olivas-Padilla2020}, and the statistics were used to infer which tasks expose operators to a higher ergonomic risk. These were assigned to the collaborative robot, and only the safer tasks in the TV assembly were left to the human operators.

\subsection{Results and discussion of the ergonomic evaluation}
Prior to segmenting professional tasks into motion primitives, the 14 HMMs were configured with a left-right topology and seven internal states. Then, their recognition performance was evaluated using the all-shots approach. The overall F-score obtained was 95.65\%, which is adequate for discriminating between motion primitives. Table \ref{table_scores} illustrates the mean, standard deviation, and mode of the ergonomic scores calculated for each task.

\begin{table}[h]
\renewcommand{\arraystretch}{1.3}
\caption{Summary statistics of the EAWS scores calculated for each task.}
\label{table_scores}
\centering
\begin{tabular}{cccc} 
\hline
\multirow{2}{*}{\textbf{Tasks}} & \multicolumn{3}{c}{\textbf{EAWS scores }}  \\ 
\cline{2-4}
                                & Mean  & STD  & Mode                        \\ 
\hline
T\textsubscript{1}                              & 16.02 & 2.65 & 17.50                       \\
T\textsubscript{2}                              & 15.02 & 3.43 & 16.00                       \\
T\textsubscript{3}                              & 10.76 & 3.68 & 8.50                        \\
T\textsubscript{4}                              & 11.50 & 3.19 & 12.50                        \\
\hline
\end{tabular}
\end{table}

According to these results, the majority of iterations of tasks T\textsubscript{1} and T\textsubscript{2} can be classified as medium-risk motions, while iterations of tasks T\textsubscript{3} and T\textsubscript{4} are classified as low-risk motions. For T\textsubscript{1}, the motion primitives detected most frequently resembled motions in which the elbows are raised above shoulder level while the torso is laterally bent. These results are expected based on the movements performed in T\textsubscript{1}, as operators must rotate and laterally bend their torsos to retrieve a circuit card from a container. Due to the container's location, operators must also raise their arms above shoulder level, as illustrated in Fig. \ref{fig:GV2a}. The motion primitives detected for T\textsubscript{2} corresponded more to motions where there is both bending and rotation of the torso and stretching of the arms.  These results match T\textsubscript{2}, where the operators bend to pick up a wire from a container and then connect it to the circuit card. The motion primitives detected for T\textsubscript{3} and T\textsubscript{4} were primarily torso rotations and working with arms bent around 90° degrees. These motion primitives are visible in Fig \ref{fig:GV2c} and \ref{fig:GV2d}. In these tasks, operators were only required to slightly rotate their torso due to the constant movement of the TV frame caused by the conveyor belt, but they did not need to stretch their arms or strongly bend forward to reach the TV frame.

Since T\textsubscript{1} and T\textsubscript{2} involve assuming awkward postures such as rotating the torso while bending forward or raising the arms above shoulder level, they represent a major ergonomic risk than T\textsubscript{3}and T\textsubscript{4}. EAWS recommends that, if possible, tasks with moderate risk must be redesigned; otherwise, the risk must be controlled through other means. Therefore, it is proposed to delegate T\textsubscript{1} and T\textsubscript{2} to the robot and leave T\textsubscript{3} and T\textsubscript{4} to the operators.

\section{optimization of work-space scenario} \label{HRCsec}
As suggested in the preceding section, it is necessary to modify the existing scenario in order to avoid operators developing musculoskeletal injuries. Thus, this section details the integration of a collaborative robot into the television manufacturing process. The framework includes a real-time gesture recognition module that enables natural HRC by performing only gestures that are convenient for the operator. Additionally, pose estimation was integrated using a skeleton-tracking algorithm to supply the robot with human pose information, allowing it to spatially adapt its movements to the operators' anthropometrics, thereby improving the operators' posture. Three experiments were conducted to determine whether gesture recognition and pose estimation can reduce the operator's range of motion. Two KPIs were utilized to evaluate the proposed HRC: the percentage of spatial adaptation achieved by the robot and a new KPI for collaborative systems presented in this paper. The new KPI quantifies the reduction in operator motion caused by the inclusion of gesture recognition.

\subsection{Gesture recognition with 3DCNNs}
3D Convolutional Neural Networks (3DCNNs), a type of Deep Learning architecture, are used for the gesture recognition part that concerns the communication between the operator and the robot. These networks were trained on a television assembly dataset generated using an RGB camera in an egocentric view. The challenge with using egocentric data for recognition is that the head and hands move in parallel but do not always follow each other. Even if the hands are a significant part of this module, they can be prominent within the frame but can also be partly or even totally out of the main view. Nevertheless, the added value of an egocentric perspective is that the network will provide gesture recognition results independent of the users' anthropometric characteristics and position in the environment.

\begin{figure}
\vspace{10pt}
\centering
{\captionsetup{position=bottom,justification=centering}
	\begin{subfigure}{0.3\linewidth}
		\centering
		\includegraphics[width=1\textwidth, height=15mm]{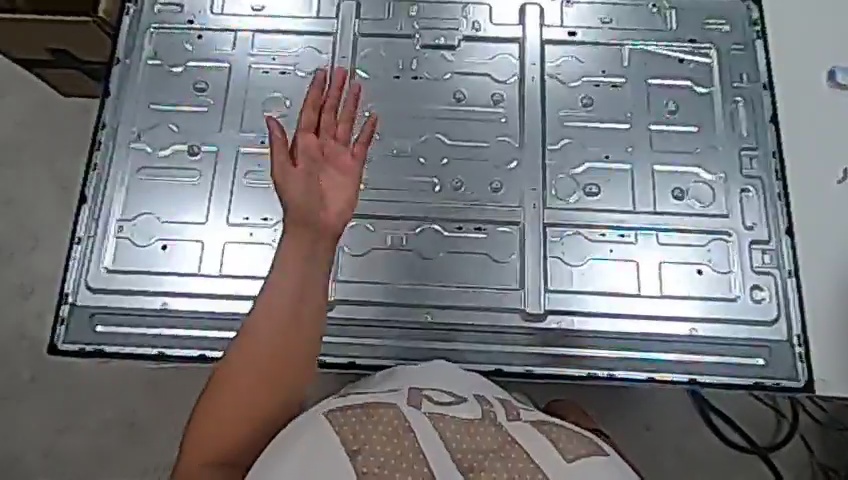} 
		\caption{G\textsubscript{1}:Start \newline}\label{fig:1}
	\end{subfigure}
	\begin{subfigure}{0.3\linewidth}
		\centering
		\includegraphics[width=1\textwidth, height=15mm]{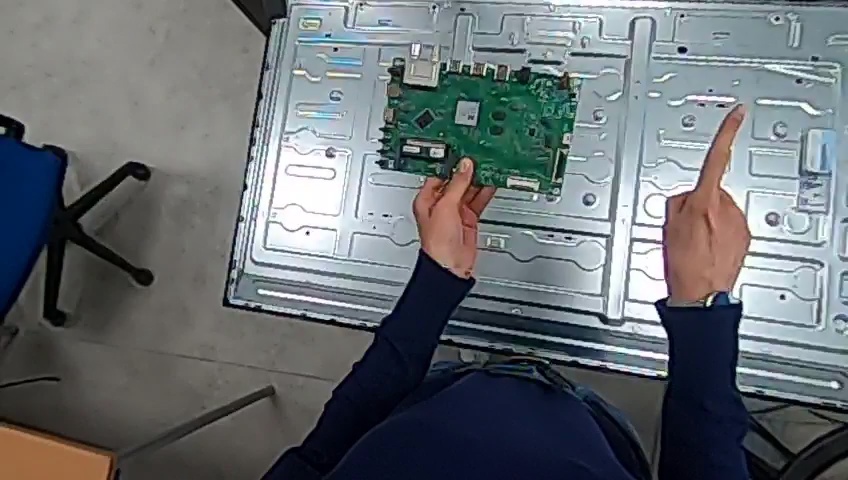}
		\caption{G\textsubscript{2}: Green card functioning}\label{fig:2}
	\end{subfigure}
	\begin{subfigure}{0.3\linewidth}
		\centering
		\includegraphics[width=1\textwidth, height=15mm]{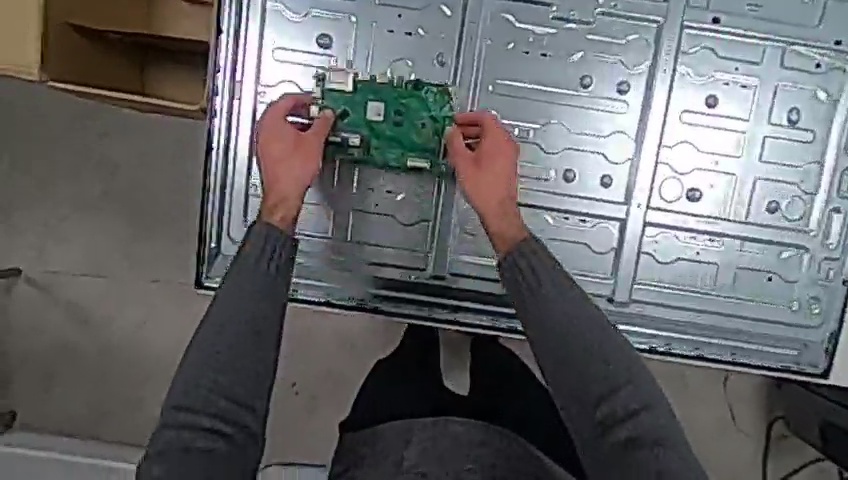}
		\caption{G\textsubscript{3}:Place green card}\label{fig:3}
	\end{subfigure}
	\begin{subfigure}{0.3\linewidth}
		\centering
		\includegraphics[width=1\textwidth, height=15mm]{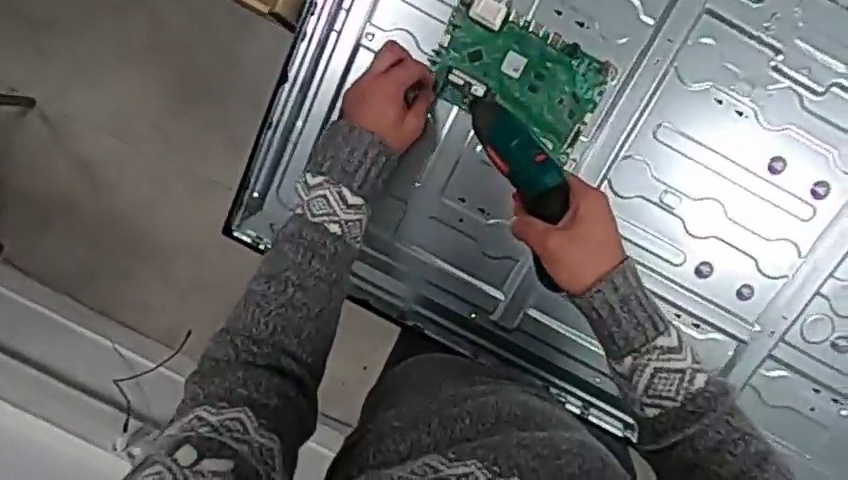}
		\caption{G\textsubscript{4}: Screw green card}\label{fig:4}
	\end{subfigure}
	\begin{subfigure}{0.3\linewidth}
		\centering
		\includegraphics[width=1\textwidth, height=15mm]{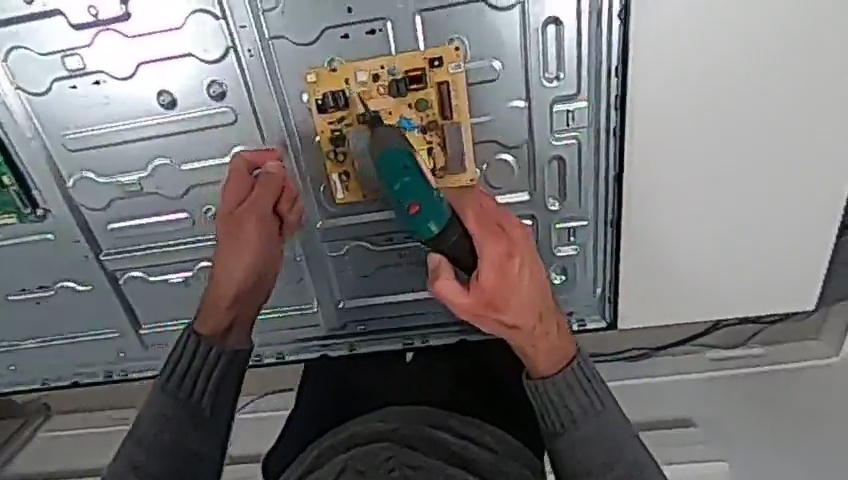}
		\caption{G\textsubscript{5}: Screw gold card}\label{fig:5}
	\end{subfigure}
	\begin{subfigure}{0.3\linewidth}
		\centering
		\includegraphics[width=1\textwidth, height=15mm]{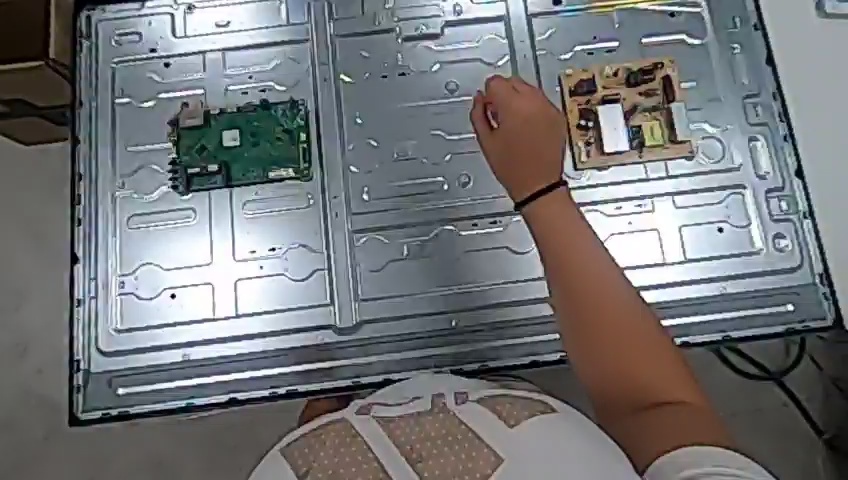}
		\caption{G\textsubscript{6}: End \newline}\label{fig:6}
	\end{subfigure}
	\begin{subfigure}{0.3\linewidth}
		\centering
		\includegraphics[width=1\textwidth, height=15mm]{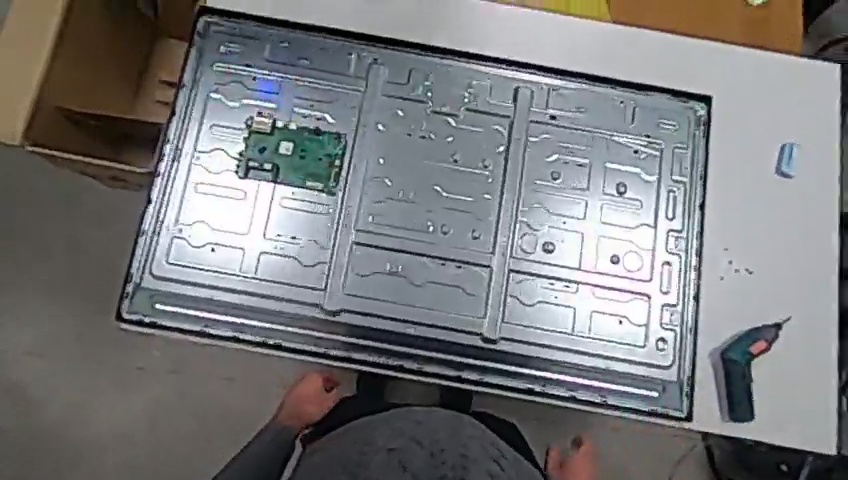}
		\caption{G\textsubscript{7}: Waiting \newline}\label{fig:7}
	\end{subfigure}
	\begin{subfigure}{0.3\linewidth}
		\centering
		\includegraphics[width=1\textwidth, height=15mm]{ 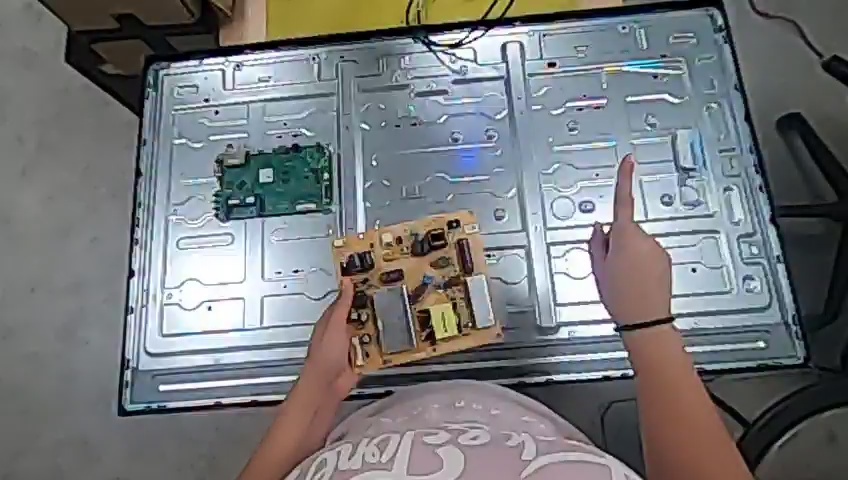}
		\caption{G\textsubscript{8}: Gold card functioning}\label{fig:8}
	\end{subfigure}
	\begin{subfigure}{0.3\linewidth}
		\centering
		\includegraphics[width=1\textwidth, height=15mm]{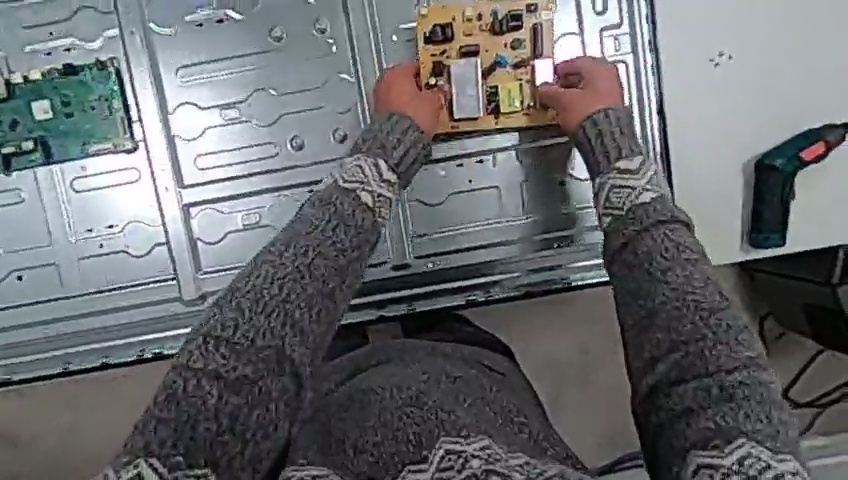}
		\caption{G\textsubscript{9}: Place gold card }\label{fig:9}
	\end{subfigure}
	\begin{subfigure}{0.3\linewidth}
		\centering
		\includegraphics[width=1\textwidth, height=15mm]{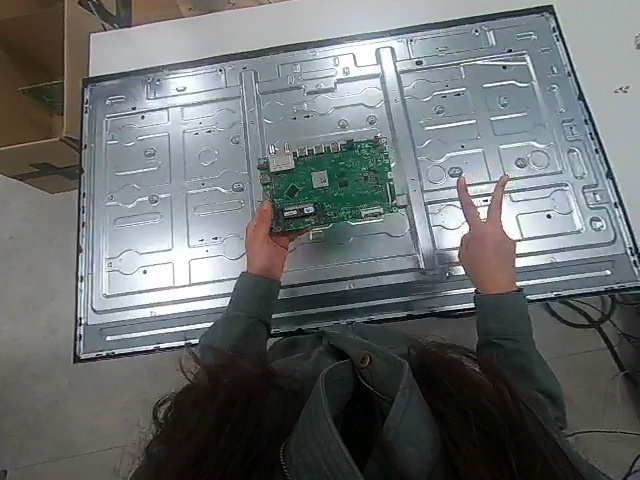}
		\caption{G\textsubscript{10}: Green card not functioning}\label{fig:10}
	\end{subfigure}
	\begin{subfigure}{0.3\linewidth}
	    \vspace{10pt}
		\centering
		\includegraphics[width=1\textwidth, height=15mm]{ 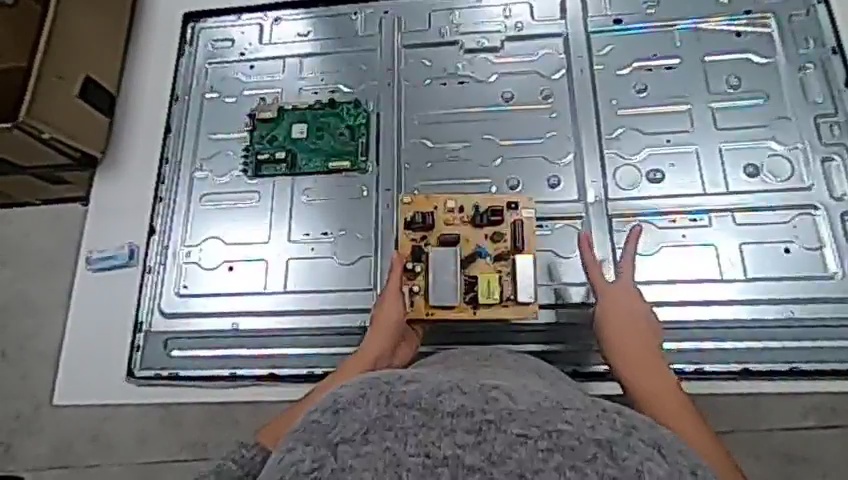}
		\caption{G\textsubscript{11}:Gold card not functioning \newline}
		\label{fig:11}
	\end{subfigure}}
	\caption{Professional gestures of the egocentric gesture recognition module.}
	\label{fig:dataset}
\end{figure}

To train the proposed network, the routine of assembling a television is divided into sub-tasks performed either by the robot or operators. The operators are assigned the ergonomically safe task that also requires decision-making.   The dataset created from this HRC scenario consists of RGB images recorded with a GoPro Black, placed with a headband on the head of the human operator, providing the egocentric view. The recording was made at an 848x480 resolution and 20 frames per second. A group of 14 operators, four females and ten males, were recorded performing six gestures and five postures. Thus, there are 11 classes in total, each corresponding to a unique command for the collaborative robot. Fig. \ref{fig:dataset} presents the commands contained in the data set. The objects involved in this TV assembly process are a TV frame and two circuit cards: the power supply (PSU-gold card) and the mainboard (chassis-green card). An earlier study \cite{Papanagiotou2021} demonstrated the network's ability to recognize the 11 gestures with a 98.5\% accuracy.

\subsection{Human Robot Collaboration for professional environments}
For the proposed HRC scenario, the UR3 robotic arm\footnote{\url{https://www.universal-robots.com/products/ur3-robot/}} from Universal Robots was used. The external parts that were used for grasping and releasing the circuit cards, were from ROBOTIQ (gripper: 2F-140\footnote{\url{https://robotiq.com/products/ft-300-force-torque-sensor}} \& force torque sensor: FT-300-S\footnote{\url{https://robotiq.com/products/2f85-140-adaptive-robot-gripper}}). For the control of the robotic arm, the Robot Operating System (ROS) was used.  
The robotic arm receives real-time ID messages from the gesture recognition module through a UDP communication protocol.
In the new routine, the operator performs the Start gesture (G\textsubscript{1}) to notify the robot that the TV assembly routine starts. The robot approaches the card container, retrieves the initial green card, and hands it to the operator in a predefined handover position. By pressing the Force sensor, the operator releases the card and verifies its functionality. If the card is functional, the operator performs G\textsubscript{2} and places it on the television frame (G\textsubscript{3}), while the robotic agent moves towards the card box to retrieve the gold card. If the green card is not functional, the operator executes G\textsubscript{10} to notify the robot, which then brings a replacement green card. When the robotic arm delivers a functional card, the operator performs G\textsubscript{2} and then places (G\textsubscript{3}) and screw (G\textsubscript{4}) the card on the television frame. This procedure is repeated until both the green and gold cards are placed on the television frame. Then, the human operator executes G\textsubscript{11} to signal the routine's completion. 

The gesture recognition module was designed with a dual-level of control to ensure the operator's safety in the HRC scenario. More precisely, the robot receives the ID of the recognized gesture only after it has been correctly recognized for 20 consecutive time frames. This may raise concerns about any latency observed during the television assembly process. However, it was observed that the time interval between capturing a frame and its correct recognition was between 0 and 0.08 ms, indicating that no significant latency was observed. The second level of control was performed within ROS, where the robot accepted only gestural IDs that corresponded to the TV assembly workflow as defined by the working routine. As a result, if an unintended incident occurs, the operator retains full control because the robotic arm strictly adheres to the routine sequence.

The introduction of the robotic arm and a gesture recognition module requires only minor torso rotations from the human operator. To facilitate natural collaboration and assist the operator in performing only ergonomically safe motions, a posture estimation module was added that enables the robot to spatially adapt to the operator. Papanagiotou et al. \cite{Papanagiotou2021} presented this posture estimation module and spatial adaptation procedure in a previous study. The spatial adaptation refers to the fact that the robotic arm does not place the cards in a fixed position but rather adapts to the operator's anthropometric characteristics. This procedure ergonomically improves the operator's posture. An Intel-RealSense RGB-D camera was utilized to capture the operator for the robot's spatial adaptation. The X and Y axes of the camera were first aligned with the X and Z axes of the robot. Then, the operator's skeleton and the position and velocity of the operator's wrists were extracted. The robotic arm tracks the operator's hand and approaches it with a circuit card when it remains motionless and in an accessible position for the robot.

\subsection{Key performance indicators}
As a key performance indicator (KPI), the percentage of robot spatial adaptation was used \cite{Papanagiotou2021}. In addition, this paper presents a second KPI for determining how much operator effort is reduced as a result of the gesture recognition module. 

The KPI of robot spatial adaptation \(SA\) represents the ratio of the distance covered by the robot without spatial adaptation to the distance covered when the robot adjusts to the operator-specified position. The following formula is used to determine this KPI:

\begin{equation} \label{eq:SA}
    SA (\%) = \frac{\|AHP - WP\| - \| PHP - WP\|}{\| PHP - WP\|}
\end{equation}\label{eqSA}

Where \(SA\) is spatial adaptation, \(AHP\) the adapted handover position, \(WP\) is the waiting point and \(PHP\) the particular handover position. Centimeters are used to measure distances. The higher the rate of adaptation, the more effort the operator had to put in during the robot collaboration.

The second KPI quantifies the difference in operators' motion before and after the introduction of gesture recognition. This KPI was calculated as follows:

\begin{equation}\label{eqRiom}
    RiOM (\%) = \frac{\|MwoGR\| - \|MwGR\|}{\|MwoGR\|}
\end{equation}

In equation \ref{eqRiom}, \(RiOM\) corresponds to the reduction in operator's motion, \(MwoGR\) the motion without gesture recognition and \(MwGR\) is the motion with gesture recognition. This KPI measures the amount of effort reduced by the operator as a result of gesture recognition.

\subsection{Experimental results}
In order to evaluate the HRC scenario in terms of collaboration and operator performance, 14 operators were recorded performing the proposed work routine in three separate experiments. In the initial experiment, gesture recognition and spatial adaptation were disabled. Therefore, operators were required to interrupt their routine and inform the robotic arm of their current action by pressing its force-torque sensor. The gesture recognition module was enabled for the second experiment, but not the spatial adaptation, so the operators received the circuit cards from a pre-defined handover position. Finally, in the third experiment, gesture recognition and spatial adaptation were enabled and used to continuously provide information about the operators' actions to the robotic arm. The purpose of these three experiments was to assess the robotic arm's adaptation and the operator's motion during each experiment.

The KPIs for each operator were determined based on the experiments and are shown in Table \ref{table_kpis}. Note that the smaller the percentage for $SA$, the better, as the robot placed the circuit cards closer to the operator's hand, reducing the necessity for the operator to reach for the card. On the other hand, for $RiOM$, the bigger the percentage is, the better because it means a greater reduction in operator motion when gesture recognition is used.

\begin{table*}
\vspace{10pt}
\renewcommand{\arraystretch}{1.3}
\caption{Measured KPIs for each operator}
\label{table_kpis}
\centering
\begin{tabular}{ccccccccccccccc} 
\hline
\multirow{2}{*}{\textbf{KPI}} & \multicolumn{14}{c}{\textbf{Operator}}                                                           \\ 
\cline{2-15}
                              & 1    & 2    & 3    & 4    & 5    & 6    & 7    & 8    & 9    & 10   & 11   & 12   & 13   & 14    \\ 
\hline
SA (\%)                       & 39.10 & 33.30 & 21.10 & 27.50 & 30.40 & 31.90 & 27.10 & 31.80 & 13.40 & 33.90 & 43.50 & 32.10 & 18.70 & 27.40  \\
\textit{RiOM (\%)}            & 31.40 & 33.10 & 24.40 & 27.10 & 32.10 & 27.30 & 24.50 & 26.80 & 37.40 & 20.60 & 45.90 & 20.80 & 21.30 & 24.50  \\
\hline
\end{tabular}
\end{table*}

\subsection{Discussion}
\begin{figure}
\vspace{10pt}
\centering
{\captionsetup{position=bottom,justification=centering}
	\begin{subfigure}{0.3\linewidth}
		\centering
		\includegraphics[width=1\textwidth, height=25mm]{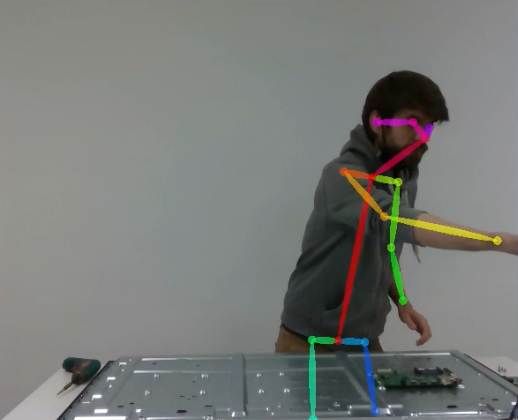}
		\caption{}\label{fig:KPIa}
	\end{subfigure}
	\begin{subfigure}{0.3\linewidth}
		\centering
		\includegraphics[width=1\textwidth, height=25mm]{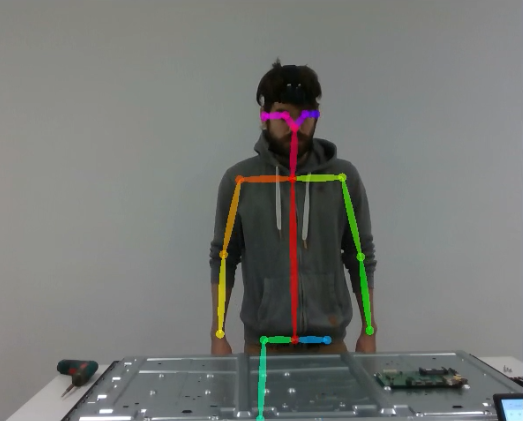}
		\caption{}\label{fig:KPIb}
	\end{subfigure}}
	\caption{Professional gestures for TV assembly with and without gesture recognition. (\textbf{a}) MwoGR: Press robot to start; (\textbf{b}) MwGR: The robot recognizes ‘Waiting gesture’ (G\textsubscript{7}) to start.}
	\label{fig:KPI}
\end{figure}

Both KPIs demonstrated a consistent decrease in the amount of physical effort required of the operator in the proposed HRC with gesture recognition and spatial adaptation. The average rate of spatial adaptation and reduction in the operators' motion among the 14 subjects was 29.37\% and 28.37\%, respectively. The optimized HRC configuration significantly decreased the operators' motion compared to the configuration without gesture recognition. This is also illustrated in Fig. \ref{fig:KPI}, where, in Fig. \ref{fig:KPIa}, which shows the HRC configuration without gesture recognition, the operator was required to rotate his torso in order to touch the robot's sensor, which is located outside the TV frame. In Fig. \ref{fig:KPIb}, where there is gesture recognition, the robot recognizes when the operator has completed his task and can proceed to the next action in the work routine.

These results provide solid evidence that the collaborative robot can adapt to human factors and that the proposed HRC scenario contributes to improved ergonomics and productivity. 

\section{Conclusion and future work} \label{concF}
This paper presents a methodology for task delegation and a human-robot collaboration framework that maximize ergonomics and production efficiency in a television co-production cell. At first, professional tasks performed on a television production line were captured using a minimal set of inertial sensors. Then, motion primitives with known ergonomic risks were detected across the entire professional tasks in order to estimate the tasks' ergonomic scores and identify the most dangerous ones. An optimized HRC was proposed, in which hazardous tasks were delegated to the collaborative robot. The HRC scenario was enhanced by applying gesture recognition and spatial adaptation that allowed the human operator to collaborate with the robot using gestures while avoiding unnecessary movements that could cause physical strain.

Future research will focus on developing a frame-by-frame risk prediction method. The system would be capable of predicting future motions and their ergonomic risk level. This type of system may be advantageous for HRC for various reasons. For example, the robot could provide preventive feedback to operators or protect them from ergonomic risks by actively assisting them in adopting safer postures before and during a task.


\section*{ACKNOWLEDGMENT}
The research leading to these results has received funding by the EU Horizon 2020 Research and Innovation Programme under grant agreement No. 820767, project CoLLaboratE. The authors like to express their gratitude to Arçelik A.S. for their support in this work.

\bibliographystyle{IEEEtran}
\bibliography{IEEEabrv,References}


\end{document}